\documentclass[11pt]{article}

\usepackage[preprint]{acl}

\usepackage{times}
\usepackage{latexsym}

\usepackage[T1]{fontenc}

\usepackage[utf8]{inputenc}

\usepackage{microtype}

\usepackage{inconsolata}

\usepackage{graphicx}
\usepackage{amsmath,hyperref}
\usepackage{amsthm}

\newtheorem{lemma}{Lemma}
\usepackage{booktabs}
\usepackage[normalem]{ulem}
\usepackage{multirow}
\usepackage{tcolorbox}
\tcbuselibrary{listings, breakable} 
\usepackage{cuted}
\usepackage{xcolor}
\usepackage{tabularx}
\usepackage{fontawesome5}
\usepackage{titletoc}
\lstdefinestyle{markdown}{
    basicstyle=\ttfamily\small,
    breaklines=true,
    frame=single,
    backgroundcolor=\color{gray!10},
    commentstyle=\color{green! 60!black},
    keywordstyle=\color{blue},
    stringstyle=\color{red}
}
\title{NL2Dashboard: A Lightweight and Controllable Framework for Generating Dashboards with LLMs}

\author{
    Boshen Shi, Kexin Yang, Yuanbo Yang, Guanguang Chang, Ce Chi, \\
    \textbf{Zhendong Wang}, \textbf{Xing Wang\thanks{~Corresponding author}}, \textbf{Junlan Feng\footnotemark[1]} \\
    \textit{Jiutian Research, China Mobile}, Beijing, China\\
    \texttt{\{shiboshen,wangxing\}@cmjt.chinamobile.com}
}

\begin{document}
\maketitle
\begin{abstract}
While Large Language Models (LLMs) have demonstrated remarkable proficiency in generating standalone charts, synthesizing comprehensive dashboards remains a formidable challenge. Existing end-to-end paradigms, which typically treat dashboard generation as a direct code generation task (e.g., raw HTML), suffer from two fundamental limitations: representation redundancy due to massive tokens spent on visual rendering, and low controllability caused by the entanglement of analytical reasoning and presentation. To address these challenges, we propose \textit{NL2Dashboard}, a lightweight framework grounded in the principle of Analysis-Presentation Decoupling. We introduce a structured intermediate representation (IR) that encapsulates the dashboard’s content, layout, and visual elements. Therefore, it confines the LLM's role to data analysis and intent translation, while offloading visual synthesis to a deterministic rendering engine. Building upon this framework, we develop a multi-agent system in which the IR-driven algorithm is instantiated as a suite of tools. Comprehensive experiments conducted with this system demonstrate that \textit{NL2Dashboard} significantly outperforms state-of-the-art baselines  across diverse domains, achieving superior visual quality, significantly higher token efficiency, and precise controllability in both generation and modification tasks.
\end{abstract}

\section{Introduction}
\label{sec:intro}
Data visualization acts as an essential tool for interpreting intrinsic patterns and features within complex data~\cite{lian2025survey, ye2024generative}. Recently, the intersection of Natural Language Processing (NLP) and visualization, i.e., NL2Vis, has received significant attention~\cite{yang2024foundation}. Powered by the reasoning and code-generation capabilities of Large Language Models (LLMs)~\cite{deepseekr1,qwen3}, current frameworks can effectively automate the translation of natural language prompts into visual results~\cite{beasley2024pipe,wang2025data,li2024prompt4vis}. This advancement has substantially reduced the reliance on user expertise, enabling the rapid generation of high-quality charts.
\begin{figure}
    \centering
    \includegraphics[width=\linewidth]{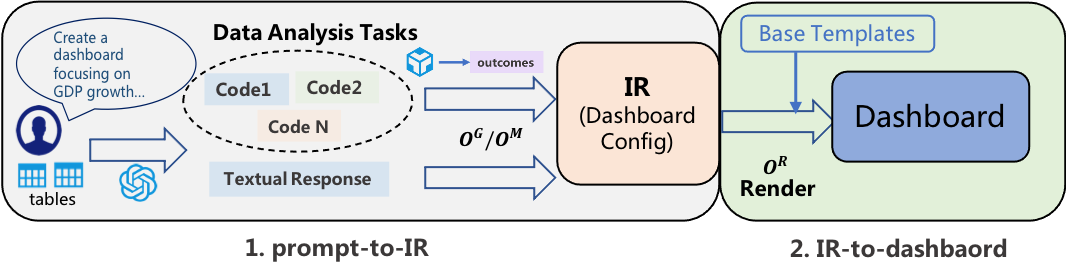}
    \caption{NL2Dashboard Framework with IR }
    \label{fig::idea}
\end{figure}

While LLMs have demonstrated remarkable proficiency in generating standalone visualizations~\cite{chi2025jt}, synthesizing comprehensive dashboards remains challenging. Unlike individual charts or infographics, dashboards are not merely collections of graphics but holistic visual systems designed for perceptual efficiency and multi-dimensional data analysis. They require diverse analytical perspectives, appropriate visual representations, and interactivity to support data-driven decision-making~\cite{sarikaya2018we}. 
As dashboards are typically rendered in HTML format, treating dashboard generation as a plain HTML generation task presents two key challenges: 
\begin{itemize} 
\item \textbf{Representation Redundancy}: A large proportion of tokens are consumed by generating descriptive code for visual rendering, such as HTML, CSS, and JavaScript, leaving only a small fraction available for reasoning and solving the underlying data analysis tasks. This severely limits the LLM’s ability to focus on complex analytical logic and results in low token efficiency. 
\item \textbf{Low controllability}: Coupling data analysis with visual rendering exposes the entire generation process to stochastic instability. Since users often need to iteratively revise existing dashboards, even localized edits risk corrupting the global dashboard structure. This lack of controllability is exacerbated by the LLM's limited capacity to interpret complex spatial layouts, making precise, intention-aligned modifications very difficult.
\end{itemize}

To overcome these limitations, we propose \textit{NL2Dashboard}, a lightweight framework grounded in the principle of Analysis-Presentation Decoupling. The core insight is that LLMs function best as "analytical engines" rather than "rendering engines." Instead of forcing the model to generate complex HTML structures, \textit{NL2Dashboard} introduces a lightweight, structured Intermediate Representation (IR) that abstracts visualization logic from implementation details. The IR-driven algorithm shifts the paradigm from direct generation to a two-stage "Reason-then-Render" workflow: 
\begin{itemize}
    \item \textbf{Prompt-to-IR}: On receiving user prompt and uploaded tables, \textit{NL2Dashboard} lets LLM focus exclusively on interpreting user intent, performing data analysis, and populating the IR. This minimizes computational overhead and maximizes adherence to analytical expectations. In the case of generating a dashboard from scratch, the IR is fully regenerated; in contrast, editing an existing dashboard results in a partial update of the IR.
    \item \textbf{IR-to-Dashboard}: \textit{NL2Dashboard} adopts a deterministic engine to map the IR to high-quality, interactive HTML-formatted dashboards via the slot-filling mechanism. We ensure the final output is professional and strictly controllable by utilizing a set of base templates with slots and rich styling presets.
\end{itemize}

By leveraging these principles, \textit{NL2Dashboard} offers three distinct advantages over existing practice. 1) Analytical Faithfulness: By separating data analysis from rendering, numerical results are derived from rigorous code execution rather than textual generation, significantly mitigating hallucination risks. 2) Token Efficiency: By offloading verbose styling codes to the rendering engine, LLM generates only the analytical part, achieving higher information density and lower latency. 3) Fine-grained Controllability: The structured IR acts as a stable interface for generation and modification. Users can iteratively refine specific dashboard components without triggering cascading layout errors common in raw code editing. Notably, \textit{NL2Dashboard} can be seamlessly integrated into any LLM-based workflow to enhance dashboard generation capabilities. To sum up, our contributions include:
\begin{enumerate}
    \item \textbf{A lightweight, IR-driven framework for end-to-end dashboard generation.} We are among the first to introduce a lightweight and controllable framework that enables the creation and iterative refinement of dashboards through natural language prompts. The dashboard is dynamic and interactive, and could support in-depth data analysis and exploration in real-world scenarios.
    \item \textbf{Agentic orchestration with executable tools.} We propose a specialized multi-agent system grounded in the proposed IR-driven algorithm, comprising a Planner, Coder, and Critic. In particular, we instantiate the IR-driven algorithm as callable tools and customize prompts to optimize the entire pipeline. Additionally, such design allows the Coder to perform rigorous code-based data analysis with self-debugging, while the Critic ensures visual fidelity, thereby guaranteeing analytical faithfulness and eliminating hallucination.
    \item \textbf{State-of-the-art performance in generation and modification.} Through comprehensive experiments across diverse domains, we demonstrate that \textit{NL2Dashboard} significantly outperforms existing end-to-end baselines. Specifically, our method achieves superior \textbf{token efficiency} and ensuring \textbf{fine-grained controllability} during complex modification tasks. The results validate that our approach offers a robust solution for trustworthy and interactive data storytelling.
\end{enumerate}

\begin{figure*}
    \centering
    \includegraphics[width=0.7\linewidth]{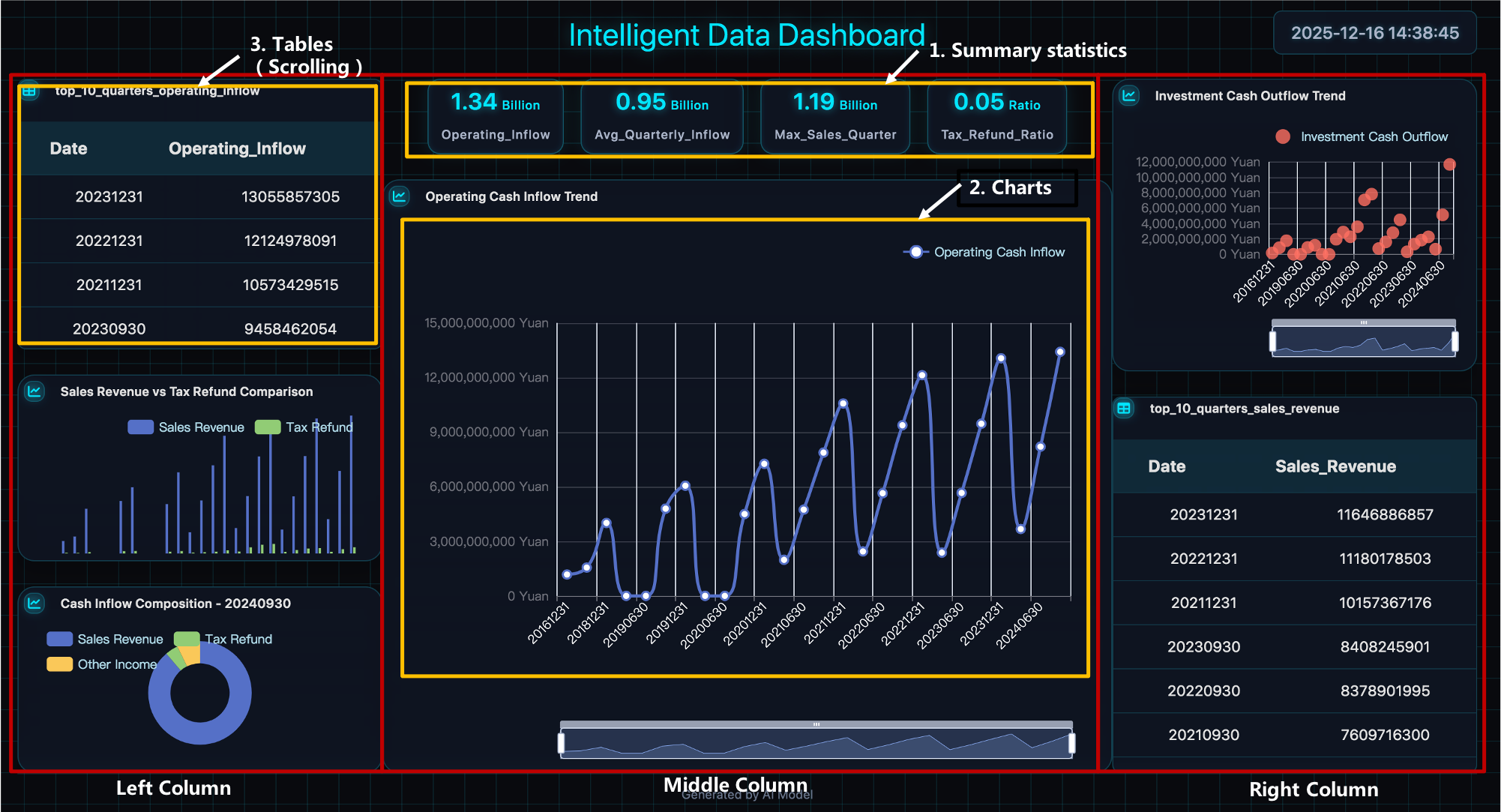}
    \caption{Example of a balanced sheet from NL2Dashboard (Dark-style Template)}
    \label{fig::example}
\end{figure*}

\section{Related Works}

To make complex data accessible to non-experts, recent work has leveraged AI for automated visualization~\cite{wu2021ai4vis,shen2022towards}. Early approaches used NLP and rule-based systems to recommend charts from queries or keywords~\cite{dibia2019data2vis,luo2018deepeye,wu2022nl2viz}, while modern methods increasingly adopt LLMs for end-to-end visualization generation~\cite{lian2025survey,ye2024generative,yang2024foundation}, including single-model~\cite{dibia2023lida,tian2024chartgpt} and multi-agent architectures~\cite{ouyang2025nvagent,li2025metal,goswami2025plotgen,chen2025coda,yang2024matplotagent}. These efforts have expanded beyond single charts to richer formats like presentations~\cite{zheng2025pptagent}, posters~\cite{zhang2025postergen}, and videos~\cite{shen2024data}.

Among these, \textbf{dashboards}, which serve as high-density, multi-view visual interfaces, remain underexplored. Pioneering works such as D2D~\cite{d2d}, Drillboards~\cite{drillboards}, and DashBot~\cite{dashbot} prioritize analytical task integration over visual quality. LADV~\cite{ladv} generates dashboards from hand-drawn sketches, limiting its applicability to natural language interaction. Most closely related is DashChat~\cite{dashchat}, which uses a multi-agent system for end-to-end NL-to-dashboard generation; however, it incurs excessive token consumption and lacks fine-grained control during modification, often introducing unintended changes. This highlights the need for a lightweight, controllable framework that faithfully translates natural language instructions into high-quality dashboards.

\section{Task Formulation}
\label{sec:formulation}
A Dashboard is defined as a five-tuple: $(S,C,T,P,B)$:
\begin{enumerate}
    \item \textbf{Analytical Components ($\{S, C, T\}$)}: Following the principles~\cite{bach2022dashboard,sarikaya2018we,bach2022dashboard}, such design space embodies the data insights. $S$ denotes summary statistics (metrics); $C$ represents visualization charts; and $T$ refers to extracted structured tables. They build the basic visualization components of a dashboard, with each focusing on a particular analysis dimension.
    \item \textbf{ Render Components ($\{P, B\}$)}: These define the presentation logic. $P$ is the dashboard config file serving as IR, it involves metadata, global layout of analytical components, and the ID of a specific base template $B$. 
\end{enumerate}
Formally, we model the dashboard generation task as a mapping from input space $\mathcal{I}=(\text{Prompt}, \text{Table})$ to a visual presentation space $\mathcal{V}$ with dashboards. Such mapping is characterized as the IR-driven algorithm containing two phases: Prompt-to-IR and IR-to-Dashboard.

\section{Methodology}
\label{sec:method}
\begin{figure*}
    \centering
    \includegraphics[width=0.8\linewidth]{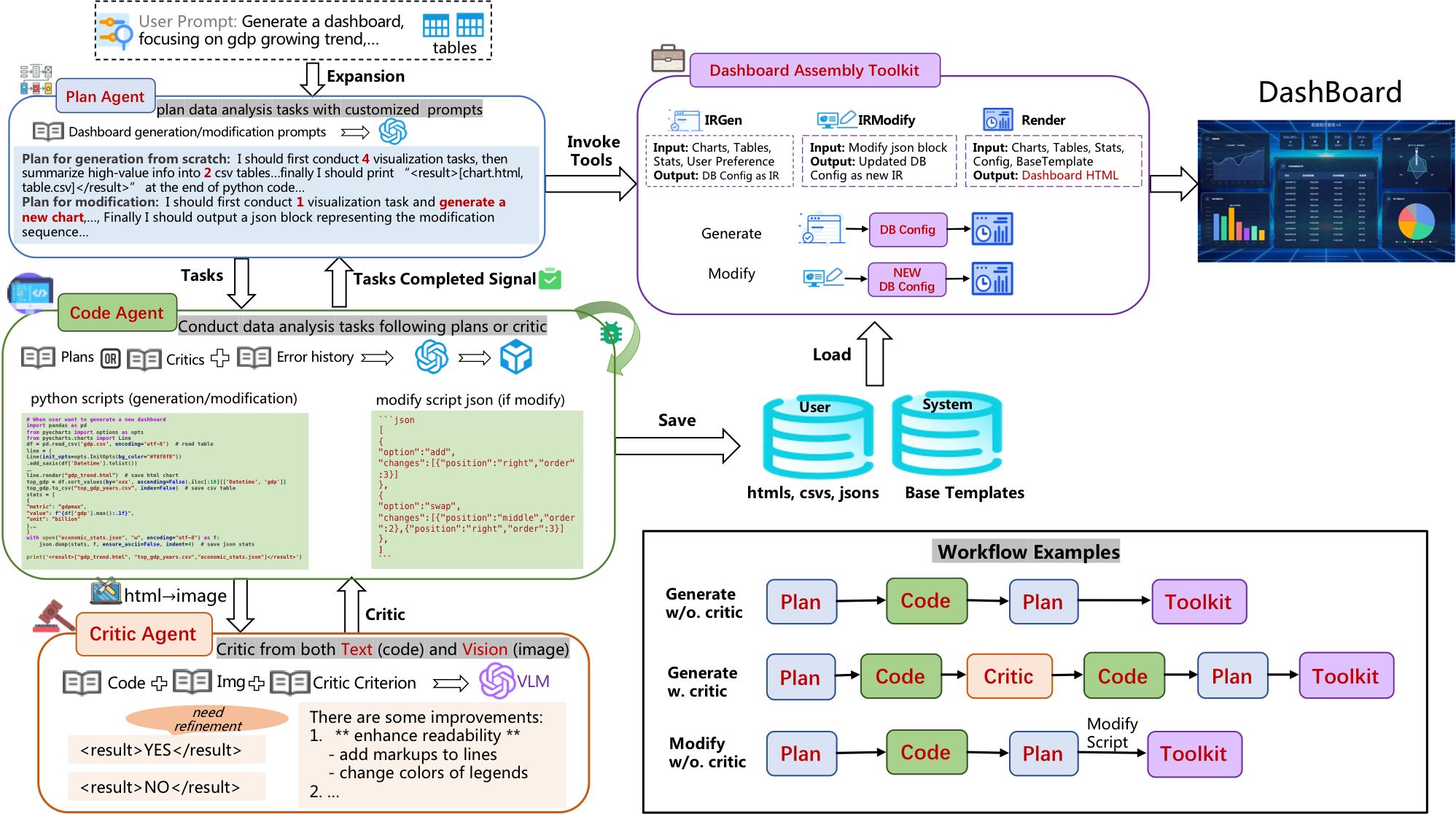}
    \caption{NL2Dashboard with a MultiAgent architecture}
    \label{fig:workflow}
\end{figure*}

In practice, users need not only to create dashboards from scratch but also to iteratively refine existing ones until they achieve a satisfactory result. Practically, we separately design the IR-driven generation algorithm and modification algorithm, which totally involve three customized operators and could be seamlessly integrated into any LLM-based workflows. In this section, we first introduce the generic generation and modification algorithm. Then, we introduce a multi-agent system implementation shown in Figure~\ref{fig:workflow}. Finally, we provide a theoretical justification for the soundness of the proposed framework.
\subsection{Dashboard Generation}
The generation algorithm focuses on creating a dashboard from scratch. Therefore, the Prompt-to-IR phase involves two steps: prompt expansion and task conducting:
\begin{itemize}
    \item \textbf{Prompt Expansion}: It first expands the original user prompt with a customized prompt for generation. This customized prompt specifies the categories of data analysis tasks, their execution order, and output specifications, guiding the model to sequentially perform a series of domain-specific tasks and store the results in a standard format. Additionally, the table schema is also injected into the user prompt to facilitate understanding, including column names, data types, and sample contents from the first few rows. 
    \item \textbf{Task Conducting}: With optimized prompts, the LLM generates and runs scripts with a sandbox to conduct several data analysis tasks, and saves the results to disks, including html ($C$), csv($T$), and json ($S$) files. 
\end{itemize}

After all tasks are completed, it generates the IR, i.e., the dashboard configuration file $P$, with a special operator $O^G$. $O^G$ selects a base template $B$ according to user preference, creates a default $P$ and populates it mainly with three types of information:
\begin{enumerate}
    \item \textbf{Default Properties}: Such properties include default dashboard settings like dashboard title, footnote, and font color.
    \item \textbf{Base Template ID}: It represents a base template, which is an HTML snippet embedded with complex CSS and JavaScript codes, while leaving content areas unpopulated. Such slots are later filled with concrete data analysis results. They are generated offline with LLM, ensuring stylistic diversity in the dashboards.
    \item \textbf{Analytical Components}: For each component, we record its file path and spatial-layout specifications. Particularly, we employ a 2D coordinate system to place each analytical component, where the x-axis is partitioned into left-middle-right or left-right segments according to the template, and the y-axis is indexed from top to bottom as 1, 2, and 3. 
\end{enumerate}

The second phase, IR-to-Dashboard, uses another deterministic operator $O^R$ to construct dashboard html from $P$ following the principle of slot filling. It first converts the contents of $S$, $C$, and $T$ into HTML fragments according to the layout specifications in $P$, then it retrieves the base template $B$ according to its ID. Finally, $O^R$ injects these fragments together with other textual fields in $P$ into the predefined slots in $B$. Putting all ingredients together, the generation algorithm could be formulated as
\begin{align*}
    S,C,T&\leftarrow LLM(\mathcal{I}) \\
    P&\leftarrow O^G(S,C,T) \\
    \text{Dashboard}&\leftarrow O^R(P;S,C,T;B)
\end{align*}

By formalizing the generation process into these distinct stages, we effectively decouple numerical reasoning from visual rendering. The \textit{Prompt-to-IR} phase ensures analytical faithfulness by grounding data insights ($S, C, T$) in the code execution and generates $P$. Subsequently, through deterministic operator $O^R$, the \textit{IR-to-Dashboard} phase guarantees that the final output strictly adheres to the design specifications. $P$ not only guarantees the renderability of the initial dashboard but also provides a stable, manipulatable handle for the subsequent modification tasks.

\subsection{Dashboard Modification}

When modifying a dashboard, LLMs typically load the entire HTML into context and regenerate the complete file, which is not only token-inefficient but also leads to uncontrolled modifications, as precise understanding, localization, and editing HTML components remain challenging. The complexity further increases when users expect to see more analysis results.

\begin{figure}
    \centering
    \includegraphics[width=\linewidth]{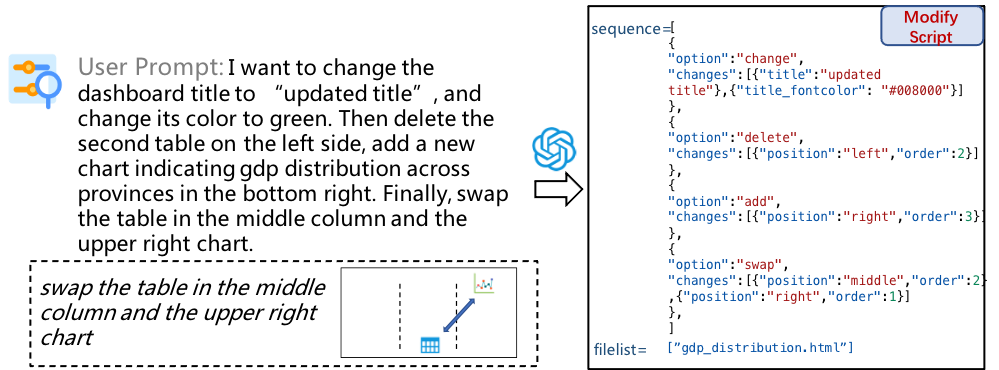}
    \caption{Translate user prompt into modify script}
    \label{fig:modify}
\end{figure}

Similar to generation, we propose an IR-driven modification algorithm to address such problems, and the main differences lie in the Prompt-to-IR phase. Therefore, a crucial problem is \textit{how to update the dashboard config with user's editing intent}. Here, we adopt an \textbf{edit-intent translation} technique, which utilizes a customized prompt to guide the LLM to first translate the user prompt into a sequence of atomic operations. By summarizing a wide range of real-world requirements, we design four atomic actions, including \textbf{change, swap, delete, and add}. The change action only changes some template-related information like background color and title, while other actions directly change the analytical components originating from $S$, $C$, and $T$. A motivating example is shown in Fig~\ref{fig:modify}, where the complex user prompt is precisely mapped to an ordered action sequence $change\rightarrow delete \rightarrow add \rightarrow swap$. The action sequence, together with the file list of newly generated analysis results (if any), is named as modify script $M$ as it defines the config update policy. Therefore, the Prompt-to-IR phase should also involve two steps: prompt expansion and task conducting:
\begin{itemize}
    \item \textbf{Prompt Expansion}: It first expands the original user prompt with a customized prompt for modification. With the optimized prompt, the LLM should 1) first output modify script $M$, and 2) plans new analysis tasks if needed. To ground the LLM in the current editing context, we encode both the prior config file and the table schema into the optimized prompt.
    \item \textbf{Task Conducting}: When new tasks are planned, the LLM generates and runs scripts with a sandbox to conduct several data analysis tasks, and saves the results to disks, including html ($C$), csv($T$), and json ($S$) files. 
\end{itemize}

After generating $M$, another operator $O^M$ is employed to update $P$. It sequentially traverses the action sequence and updates the config based on each action and its associated file (if any). Once $P$ is updated, the IR-to-Dashboard phase uses $O^R$ to re-construct the dashboard html from $P$. Putting all ingredients together, the modification algorithm could be formulated as

\begin{align*}
    M,[S',C',T']&\leftarrow LLM(\mathcal{I}) \\
    P&\leftarrow O^M(P;M;[S',C',T']) \\
    \text{Dashboard}&\leftarrow O^R(P;S\cup S',C\cup C',T\cup T';B)
\end{align*}

The modification algorithm makes editing dashboard efficient and controllable, as LLMs focus on translating user intents and planning new tasks, without regenerating the entire HTML file. Besides, the edit-intent translation technique and operator $O^M$ ensures that the editing intent is precisely identified and accurately reflected in the dashboard without affecting any unspecified components.

\subsection{Implementation with Multi-Agent System}
Both the generation and modification algorithms could be seamlessly integrated into LLM-based workflows. Here, we design a multi-agent system and instantiate the key algorithms in \textit{NL2Dashboard} as a set of callable tools. As shown in Fig~\ref{fig:modify}, the architecture involves four key components:
\begin{enumerate}
    \item Planner: It serves as the strategic controller and owns key functions. 1) Detect intent: Classifying the user prompt as dashboard generation or modification. 2) Expand prompt: Optimizing the raw user prompt with customized prompts. 3) Schedule task: Sequentially submitting tasks to the coder and checking the completion of every task. 4) Invoke tools: Invoking dashboard assembly toolkit to create dashboard once all tasks and LLM completion are finished.
    \item Coder: After receiving subtasks from planner or feedback from critic, it generates runnable scripts and executes them with a sandbox to produce analytical artifacts ($S, C, T$). The inherent self-debugging mechanism enables LLM to receive error feedback upon code execution failure and regenerate the code, thereby significantly improving the success rate of task execution.
    \item Critic: Powered by the vision-language model (VLM), it evaluates each chart ($C$) along visual dimensions. If a particular chart needs refinement, it provides detailed feedback to Coder for improvement.
    \item Dashboard assembly toolkit: We encapsulate the theoretical operators defined above as distinct, callable tools for the agents. Specifically, $O^G$, $O^R$, and $O^M$ are instantiated as \texttt{IRGen}, \texttt{DBCompile}, and \texttt{IRModify}, respectively (see Fig.~\ref{fig:workflow}).
\end{enumerate}

By orchestrating these entities, the Agents handle the reasoning and data analysis, while the Toolkit focuses on the dashboard. This collaboration ensures both the \textbf{analytical accuracy} of the content and the \textbf{fine-grained controllability} of the presentation.

\subsection{Theoretical Analysis}
\label{sec:theory}
We formalize the reliability of dashboard generation using entropy decomposition. Following the principle of analysis-presentation decoupling, let $H_{ir}$ denote the uncertainty related to generating tokens for constructing IR, including codes and $M$. Let $H_{vis}$ represent the uncertainty related to generating tokens for visual components, which is the "noise" relative to the user's analytical intent. Let $H(Y)$ be the total entropy of the dashboard sequence, so
\begin{equation*}
    H(Y) = \underbrace{H_{ir}(Y)}_{\text{Data Analysis}} + \underbrace{H_{vis}(Y)}_{\text{Visual Presentation}}
\end{equation*}

Derived from the definition of Perplexity and the Chain Rule of probability, the reliability $P_{succ} \propto e^{-H(Y)}$, so minimizing $H(Y)$ is critical. \textit{NL2Dashboard} achieves $H_{vis} \approx 0$ by utilizing base templates and 
deterministic render engine, whereas end-to-end methods struggle with $H_{vis} \gg H_{ir}$. This proves $P_{succ}(\text{\textit{NL2Dashboard}}) > P_{succ}(\text{Baselines})$. Theoretical details are provided in Appendix.

\begin{table*}
\centering
\caption{Dashboard Quality Study with LLM-As-Judge}
\resizebox{0.8\linewidth}{!}{%
\begin{tabular}{cccccc} 
\toprule
                                   &               & Insightfulness & Visual Fidelity & Information Richness & Total Score      \\ 
\midrule
\multirow{4}{*}{Generation Task}   & Doubao        & 2.96~          & 3.33~           & 3.48~                & 9.78~            \\
                                   & Gemini2.5 pro  & 2.93~          & \textbf{4.15~}  & \uline{3.56~}        & \uline{10.63~}   \\
                                   & GPT5(Agentic) & \textbf{3.22~} & 3.78~           & 3.52~                & 10.52~           \\
                                   & \textit{NL2Dashboard}          & \uline{3.04~}  & \uline{4.11~}   & \textbf{4.74~}       & \textbf{11.89~}  \\ 
\hline
\multirow{4}{*}{Modification Task} & Doubao        & 2.74~          & 3.13~           & 3.13~                & 8.99~            \\
                                   & Gemini2.5 pro & 3.04~          & \textbf{4.16~}  & \uline{3.65~}        & \uline{10.84~}   \\
                                   & GPT5(Agentic)         & \uline{3.16~}  & \uline{4.13~}   & 3.40~                & 10.69~           \\
                                   & \textit{NL2Dashboard}           & \textbf{3.20~} & 3.93~           & \textbf{4.80~}       & \textbf{11.93~}  \\
\bottomrule
\end{tabular}
}
\label{tab::quality}
\end{table*}
\begin{table*}
\centering
\caption{Token Efficiency Study w. GOR (lower is better)}
\resizebox{0.5\linewidth}{!}{%
\begin{tabular}{ccccccccc} 
\toprule
              & G             & M1             & M2             & M3             & M4             & M5             & M6             & M7              \\ 
\midrule
Doubao        & 1.59~          & 1.11~          & 2.23~          & 1.26~          & 1.12~          & 1.38~          & 2.05~          & 1.60~           \\
Gemini2.5 pro & \uline{1.00~}  & \uline{1.00~}  & \uline{1.00~}  & 1.00~          & \uline{1.00~}  & \uline{1.00~}  & \uline{1.00~}  & \uline{1.00~}   \\
GPT5(Agentic) & 2.24~          & 1.18~          & 1.52~          & \uline{0.96}~  & 1.79~          & 1.64~          & 3.30~          & 2.88~           \\
\textit{NL2Dashboard}           & \textbf{0.58}~ & \textbf{0.02~} & \textbf{0.04~} & \textbf{0.03~} & \textbf{0.32~} & \textbf{0.20~} & \textbf{0.43~} & \textbf{0.22}~  \\
\bottomrule
\end{tabular}
}
\label{tab::gor}
\end{table*}
\section{Experiment}
\label{sec:exp}
\subsection{Experimental Settings}
\textbf{Generation Task}: We first build a dataset with ten tables collected from real-world scenarios, including finance, education, and government domains. We prompt the model with a table to generate an HTML-formatted dashboard.  

\noindent \textbf{Modification Task}: For each table, we defined seven cases (M1–M7), ranging from single-step edits (e.g., title change, chart replacement; M1–M4) to multi-step sequences (e.g., swap charts then add one; M5–M7). For each case, we prompt the model with a table and the originally generated dashboard file to generate a new dashboard. 

With the above tasks, we conducted experimental studies to address:
\begin{enumerate}
    \item \textbf{RQ1}: How high is the dashboard quality?
    \item \textbf{RQ2}: How faithfully does \textit{NL2Dashboard} execute user-specified modifications?
    \item \textbf{RQ3}:  What is the token efficiency?
    \item \textbf{RQ4}: How effective is the critic-based iterative optimization mechanism in the multi-agent system?
\end{enumerate}

\noindent \textbf{Baselines}: Our baselines cover three widely-used LLM products with demonstrated capabilities in long-context processing or code-integrated reasoning: Doubao, Gemini 2.5 pro, and GPT5 with agentic mode, and we conducted evaluations using their official web interfaces directly. For \textit{NL2Dashboard}, we directly utilized the APIs provided by the Qwen3-MAX and Qwen3-VL-Plus.  

\noindent \textbf{Metrics}: We assess different frameworks along three dimensions:
\begin{enumerate}
    \item Quality: We adopted a vision-language model to evaluate rendered dashboards on \textbf{insightfulness} (the depth of data analysis), \textbf{visual fidelity} (visual clarity, layout, and rendering stability), and \textbf{information richness} (multi-dimensional coverage), each scored on a scale of 1–5.
    \item Token Efficiency: We introduced \textbf{Generative Overhead Ratio (GOR)} to calculate the token efficiency, which is defined as $\frac{\#Token_{llm}}{\#Token_{db}}$ where $\#Token_{llm}$ and $\#Token_{db}$ denote the token counts of the LLM output and the dashboard file, respectively. A smaller GOR indicates that the model can generate the dashboard with a lower token budget. Since dashboards generated by different models vary in complexity, we used GOR instead of absolute token counts. 
    \item Controllability: We calculated the \textbf{success rate (SR)} of each model in performing the modification tasks. All modification outcomes were validated with a combination of manual labeling and cross-validation through three human experts.
\end{enumerate}

\begin{figure}
    \centering
    \includegraphics[width=\linewidth]{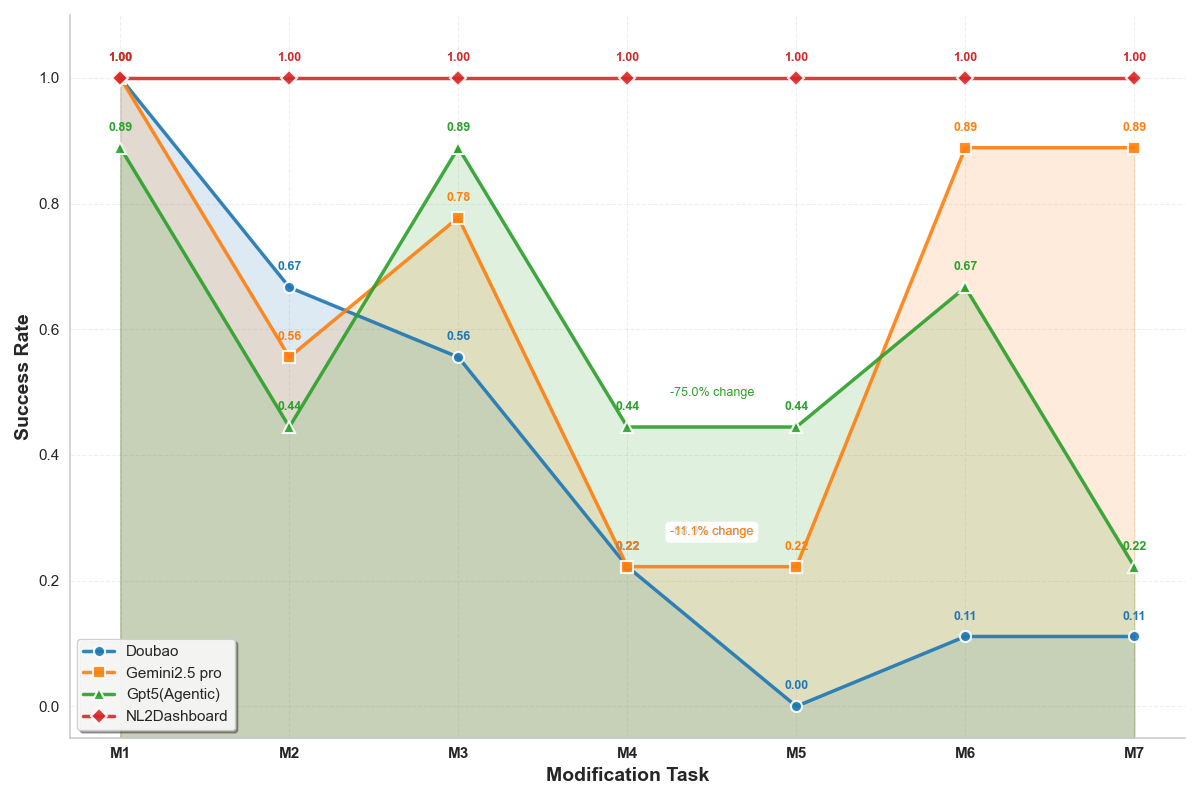}
    \caption{Modification SR with difficulty changing}
    \label{fig::mod_sr}
\end{figure}

\begin{figure*}
    \centering
    \includegraphics[width=\linewidth]{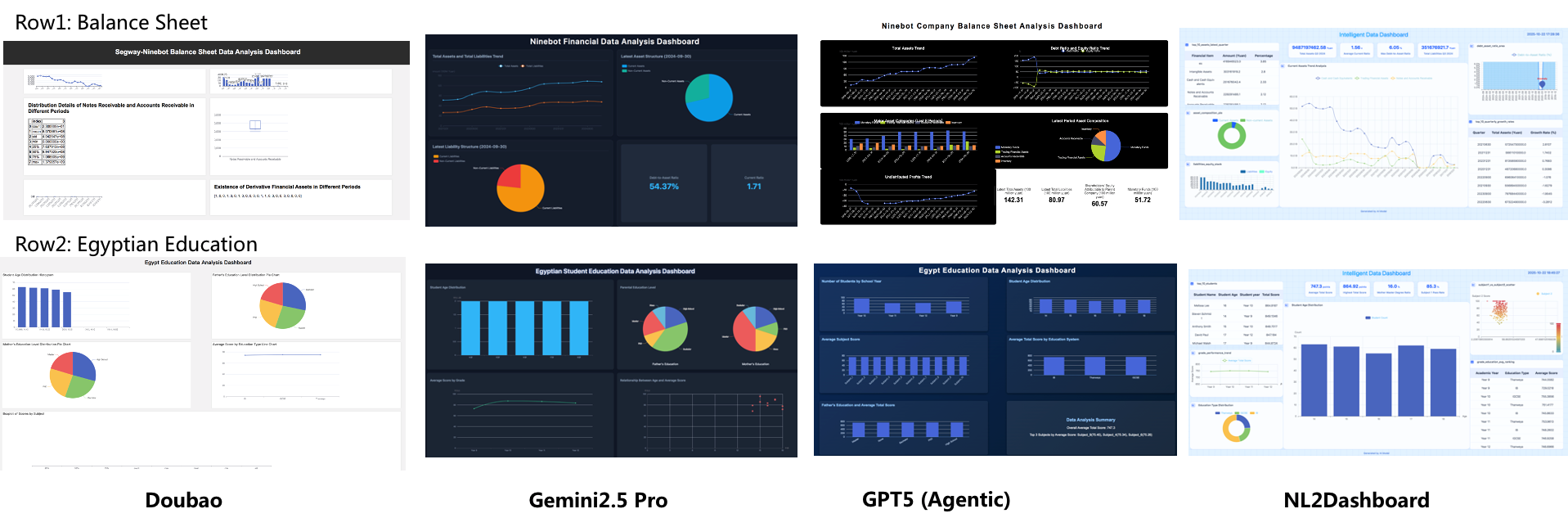}
    \caption{Comparision between models}
    \label{fig::db_compare}
\end{figure*}
\subsection{Overall Effectiveness}
To answer \textbf{RQ1}, we evaluated from the perspective of quality and summarize the results in Table~\ref{tab::quality} in both generation and modification cases. Overall, \textit{NL2Dashboard} achieves the highest quality scores and ranks within the top two in all evaluation metrics. Compared to the second-best baseline, Gemini2.5 pro, it achieves 8.4\% and 7.3\% performance improvements in generation and modification cases, respectively. 
Notably, \textit{NL2Dashboard} achieves a significant improvement on information richness, indicating that the created dashboards exhibit higher information density and greater practical utility. Notably, the proposed modification workflow does not degrade dashboard quality. In contrast, dashboards modified by baselines, like Doubao, exhibited $\sim$10\% average degradation across multiple evaluation dimensions. 

To answer \textbf{RQ2}, we further evaluated the modification success rate on the editing tasks. As shown in Fig~\ref{fig::mod_sr}, \textit{NL2Dashboard} precisely completes all tasks, outperforming baselines by 35\%–62\%. Moreover, as the task complexity increases progressively from M1 to M7, baseline success rates decline steadily due to difficulties in (1) interpreting user intents into executable operations and (2) locating target components in HTML. In contrast, our model leverages a designed modify script and an IR update operator to accurately map user intents to dashboard edits, and employs a deterministic rendering mechanism to generate the updated dashboard with high fidelity.

\subsection{Efficiency Study}
To anwer \textbf{RQ3}, we recorded the token consumption and calculate GOR and report the results in Table~\ref{tab::gor}. First, only Gemini 2.5 Pro generates dashboards by directly writing HTML code, while all other models do so by generating Python scripts and executing in a sandbox. As a result, Gemini 2.5 Pro achieves a GOR of 1. In contrast, \textit{NL2Dashboard} exhibits a GOR significantly below 1, whereas Doubao and GPT5 generally fall within the range of 1–3. Combined with the findings from \textbf{RQ1}, this indicates that \textit{NL2Dashboard} can produce high-quality dashboards with minimal token overhead. This advantage persists as task complexity grows. An exception is Task M6, which incurs higher token usage than M7 despite fewer editing steps, as it involves multiple new analytical tasks (vs. only one new task in M7).

\begin{figure}
    \centering
    \includegraphics[width=\linewidth]{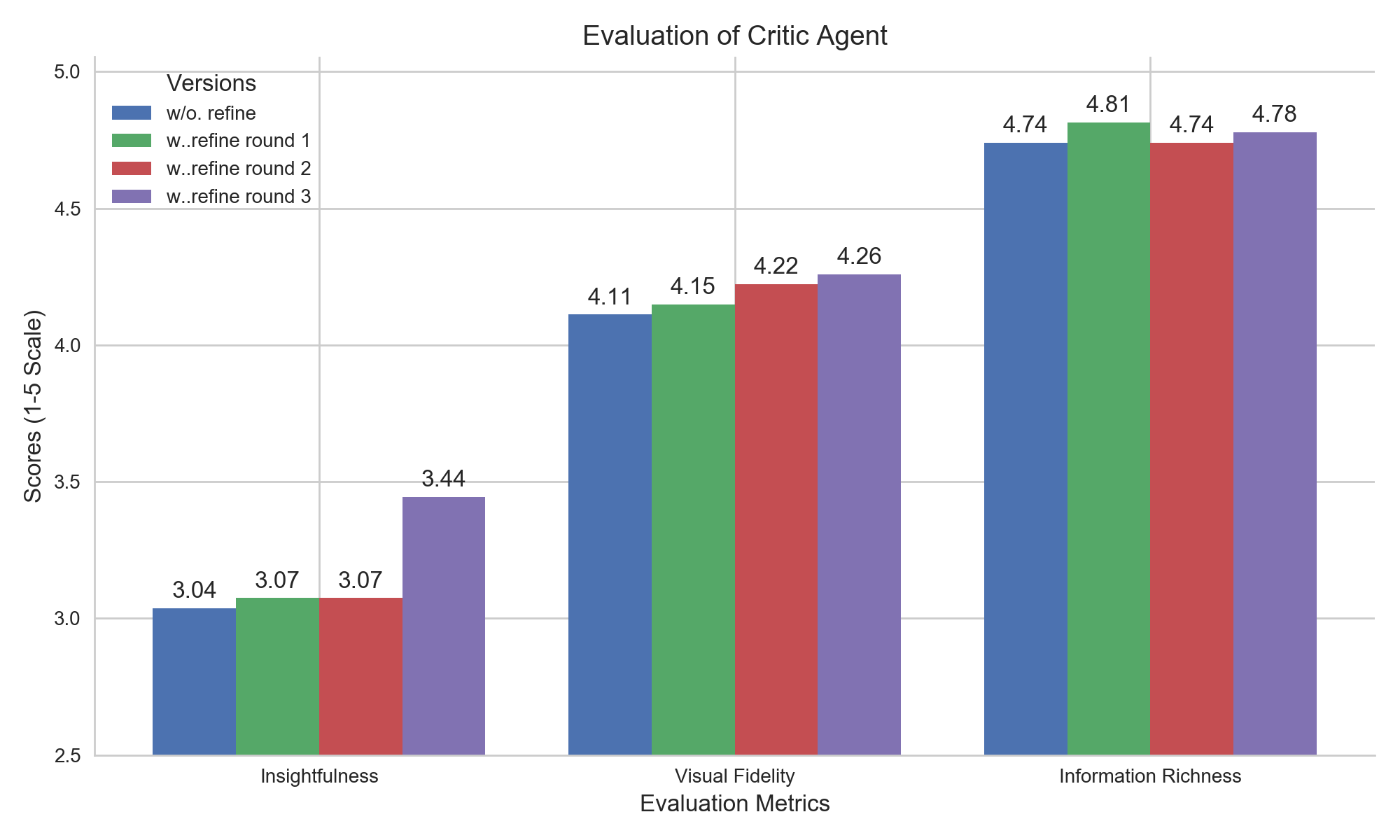}
    \caption{Ablation}
    \label{fig::ablation}
\end{figure}

\subsection{Ablation Study}
To investigate \textbf{RQ4}, we examined whether the critic agent improves dashboard quality beyond the base system (Table~\ref{tab::quality} reports final scores without critic feedback). We allowed up to three optimization rounds and track metric evolution. As shown in Fig.~\ref{fig::ablation}, all quality dimensions benefit from the critic. However, additional rounds incur diminishing returns and higher token costs. Thus, zero or one round is sufficient in practice.

\subsection{Case Study}
\label{sect::case_study}
\textbf{Generation}: As shown in Fig.~\ref{fig::db_compare}, NL2Dashboard demonstrates superior performance in these representative cases with optimized structural layout and richer visual components. In contrast, baselines suffer from sparse utilization of space, over-cluttered arrangements, or over-reliance on basic chart types (e.g., bar charts) .

\noindent \textbf{Modification}: We conducted quantitative analysis on bad cases over 210 modification tasks with baselines. The dominant failure mode is \textit{spatial reasoning} (41\%)—models struggle to manipulate coordinates and relative positions, especially during component swaps. Secondary issues include \textit{instruction adherence (specifically task omission)} (22\%) and \textit{boundary control (typically over-deletion)} (18\%). These reflect instability in layout understanding and complex instruction parsing. \textit{NL2Dashboard} mitigates these problems by encoding spatial priors into IRs and decomposing user editing intents into atomic, executable operations.


\section{Conclusion}
We present \textit{NL2Dashboard}, a lightweight and controllable framework that generates dashboards with user prompts. The core insight is decoupling data analysis from visual rendering through a structured Intermediate Representation (IR). Through theoretical analysis and comprehensive experiments over a multi-agent system implementation, we validate that \textit{NL2Dashboard} significantly minimizes generation entropy, yielding superior token efficiency, analytical faithfulness, and fine-grained controllability. We hope this work establishes a foundation for trustworthy and interactive data storytelling, paving the way for future explorations in multi-modal visual analytics.

\bibliography{custom}

\begin{thebibliography}{31}
\providecommand{\natexlab}[1]{#1}

\bibitem[{Bach et~al.(2022)Bach, Freeman, Abdul-Rahman, Turkay, Khan, Fan, and Chen}]{bach2022dashboard}
Benjamin Bach, Euan Freeman, Alfie Abdul-Rahman, Cagatay Turkay, Saiful Khan, Yulei Fan, and Min Chen. 2022.
\newblock Dashboard design patterns.
\newblock \emph{IEEE transactions on visualization and computer graphics}, 29(1):342--352.

\bibitem[{Beasley and Abouzied(2024)}]{beasley2024pipe}
Cole Beasley and Azza Abouzied. 2024.
\newblock Pipe (line) dreams: Fully automated end-to-end analysis and visualization.
\newblock In \emph{Proceedings of the 2024 Workshop on Human-In-the-Loop Data Analytics}, pages 1--7.

\bibitem[{Chen et~al.(2025)Chen, Chen, Arik, Sra, Pfister, and Yoon}]{chen2025coda}
Zichen Chen, Jiefeng Chen, Sercan~{\"O} Arik, Misha Sra, Tomas Pfister, and Jinsung Yoon. 2025.
\newblock Coda: Agentic systems for collaborative data visualization.
\newblock \emph{arXiv preprint arXiv:2510.03194}.

\bibitem[{Chi et~al.(2025)Chi, Wang, Wang, Liu, Li, Song, Zhao, Yang, Shi, Yang et~al.}]{chi2025jt}
Ce~Chi, Xing Wang, Zhendong Wang, Xiaofan Liu, Ce~Li, Zhiyan Song, Chen Zhao, Kexin Yang, Boshen Shi, Jingjing Yang, and 1 others. 2025.
\newblock Jt-da: Enhancing data analysis with tool-integrated table reasoning large language models.
\newblock \emph{arXiv preprint arXiv:2512.06859}.

\bibitem[{Deng et~al.(2022)Deng, Wu, Qu, and Wu}]{dashbot}
Dazhen Deng, Aoyu Wu, Huamin Qu, and Yingcai Wu. 2022.
\newblock Dashbot: Insight-driven dashboard generation based on deep reinforcement learning.
\newblock \emph{IEEE Transactions on Visualization and Computer Graphics}, 29(1):690--700.

\bibitem[{Dibia(2023)}]{dibia2023lida}
Victor Dibia. 2023.
\newblock Lida: A tool for automatic generation of grammar-agnostic visualizations and infographics using large language models.
\newblock In \emph{Proceedings of the 61st Annual Meeting of the Association for Computational Linguistics (Volume 3: System Demonstrations)}, pages 113--126.

\bibitem[{Dibia and Demiralp(2019)}]{dibia2019data2vis}
Victor Dibia and {\c{C}}a{\u{g}}atay Demiralp. 2019.
\newblock Data2vis: Automatic generation of data visualizations using sequence-to-sequence recurrent neural networks.
\newblock \emph{IEEE computer graphics and applications}, 39(5):33--46.

\bibitem[{Goswami et~al.(2025)Goswami, Mathur, Rossi, and Dernoncourt}]{goswami2025plotgen}
Kanika Goswami, Puneet Mathur, Ryan Rossi, and Franck Dernoncourt. 2025.
\newblock Plotgen: Multi-agent llm-based scientific data visualization via multimodal retrieval feedback.
\newblock In \emph{Companion Proceedings of the ACM on Web Conference 2025}, pages 1672--1676.

\bibitem[{Guo et~al.(2025)Guo, Yang, Zhang, Song, Wang, Zhu, Xu, Zhang, Ma, Bi et~al.}]{deepseekr1}
Daya Guo, Dejian Yang, Haowei Zhang, Junxiao Song, Peiyi Wang, Qihao Zhu, Runxin Xu, Ruoyu Zhang, Shirong Ma, Xiao Bi, and 1 others. 2025.
\newblock Deepseek-r1 incentivizes reasoning in llms through reinforcement learning.
\newblock \emph{Nature}, 645(8081):633--638.

\bibitem[{Li et~al.(2025)Li, Wang, Gu, Chang, and Peng}]{li2025metal}
Bingxuan Li, Yiwei Wang, Jiuxiang Gu, Kai-Wei Chang, and Nanyun Peng. 2025.
\newblock Metal: A multi-agent framework for chart generation with test-time scaling.
\newblock \emph{arXiv preprint arXiv:2502.17651}.

\bibitem[{Li et~al.(2024)Li, Chen, Song, Song, and Zhang}]{li2024prompt4vis}
Shuaimin Li, Xuanang Chen, Yuanfeng Song, Yunze Song, and Chen Zhang. 2024.
\newblock Prompt4vis: Prompting large language models with example mining and schema filtering for tabular data visualization.
\newblock \emph{arXiv preprint arXiv:2402.07909}.

\bibitem[{Lian et~al.(2025)Lian, Hao, Zeng, and Luo}]{lian2025survey}
Yijie Lian, Jianing Hao, Wei Zeng, and Qiong Luo. 2025.
\newblock A survey of visual insight mining: Connecting data and insights via visualization.
\newblock \emph{Visual Informatics}, page 100271.

\bibitem[{Luo et~al.(2018)Luo, Qin, Tang, Li, and Wang}]{luo2018deepeye}
Yuyu Luo, Xuedi Qin, Nan Tang, Guoliang Li, and Xinran Wang. 2018.
\newblock Deepeye: Creating good data visualizations by keyword search.
\newblock In \emph{Proceedings of the 2018 International Conference on Management of Data}, pages 1733--1736.

\bibitem[{Ma et~al.(2020)Ma, Mei, Guan, Huang, Zhang, Xin, Dai, Wen, and Chen}]{ladv}
Ruixian Ma, Honghui Mei, Huihua Guan, Wei Huang, Fan Zhang, Chengye Xin, Wenzhuo Dai, Xiao Wen, and Wei Chen. 2020.
\newblock Ladv: Deep learning assisted authoring of dashboard visualizations from images and sketches.
\newblock \emph{IEEE Transactions on Visualization and Computer Graphics}, 27(9):3717--3732.

\bibitem[{Ouyang et~al.(2025)Ouyang, Chen, Nie, Gui, Wan, Zhang, and Chen}]{ouyang2025nvagent}
Geliang Ouyang, Jingyao Chen, Zhihe Nie, Yi~Gui, Yao Wan, Hongyu Zhang, and Dongping Chen. 2025.
\newblock nvagent: Automated data visualization from natural language via collaborative agent workflow.
\newblock \emph{arXiv preprint arXiv:2502.05036}.

\bibitem[{Sarikaya et~al.(2018)Sarikaya, Correll, Bartram, Tory, and Fisher}]{sarikaya2018we}
Alper Sarikaya, Michael Correll, Lyn Bartram, Melanie Tory, and Danyel Fisher. 2018.
\newblock What do we talk about when we talk about dashboards?
\newblock \emph{IEEE transactions on visualization and computer graphics}, 25(1):682--692.

\bibitem[{Shen et~al.(2024)Shen, Li, Wang, and Qu}]{shen2024data}
Leixian Shen, Haotian Li, Yun Wang, and Huamin Qu. 2024.
\newblock From data to story: Towards automatic animated data video creation with llm-based multi-agent systems.
\newblock In \emph{2024 IEEE VIS Workshop on Data Storytelling in an Era of Generative AI (GEN4DS)}, pages 20--27. IEEE.

\bibitem[{Shen et~al.(2022)Shen, Shen, Luo, Yang, Hu, Zhang, Tai, and Wang}]{shen2022towards}
Leixian Shen, Enya Shen, Yuyu Luo, Xiaocong Yang, Xuming Hu, Xiongshuai Zhang, Zhiwei Tai, and Jianmin Wang. 2022.
\newblock Towards natural language interfaces for data visualization: A survey.
\newblock \emph{IEEE transactions on visualization and computer graphics}, 29(6):3121--3144.

\bibitem[{Shen et~al.(2025)Shen, Lin, Liu, Xin, Dai, Chen, Wen, and Lan}]{dashchat}
Siqi Shen, Ziyue Lin, Wanchen Liu, Chengye Xin, Wenzhuo Dai, Siming Chen, Xiao Wen, and Xingyu Lan. 2025.
\newblock Dashchat: Interactive authoring of industrial dashboard design prototypes through conversation with llm-powered agents.
\newblock \emph{arXiv preprint arXiv:2504.12865}.

\bibitem[{Shin et~al.(2025)Shin, Na, and Elmqvist}]{drillboards}
Sungbok Shin, Inyoup Na, and Niklas Elmqvist. 2025.
\newblock Drillboards: Adaptive visualization dashboards for dynamic personalization of visualization experiences.
\newblock \emph{IEEE Transactions on Visualization and Computer Graphics}.

\bibitem[{Tian et~al.(2024)Tian, Cui, Deng, Yi, Yang, Zhang, and Wu}]{tian2024chartgpt}
Yuan Tian, Weiwei Cui, Dazhen Deng, Xinjing Yi, Yurun Yang, Haidong Zhang, and Yingcai Wu. 2024.
\newblock Chartgpt: Leveraging llms to generate charts from abstract natural language.
\newblock \emph{IEEE Transactions on Visualization and Computer Graphics}, 31(3):1731--1745.

\bibitem[{Wang et~al.(2025)Wang, Lee, Drucker, Marshall, and Gao}]{wang2025data}
Chenglong Wang, Bongshin Lee, Steven~M Drucker, Dan Marshall, and Jianfeng Gao. 2025.
\newblock Data formulator 2: Iterative creation of data visualizations, with ai transforming data along the way.
\newblock In \emph{Proceedings of the 2025 CHI Conference on Human Factors in Computing Systems}, pages 1--17.

\bibitem[{Wu et~al.(2021)Wu, Wang, Shu, Moritz, Cui, Zhang, Zhang, and Qu}]{wu2021ai4vis}
Aoyu Wu, Yun Wang, Xinhuan Shu, Dominik Moritz, Weiwei Cui, Haidong Zhang, Dongmei Zhang, and Huamin Qu. 2021.
\newblock Ai4vis: Survey on artificial intelligence approaches for data visualization.
\newblock \emph{IEEE Transactions on Visualization and Computer Graphics}, 28(12):5049--5070.

\bibitem[{Wu et~al.(2022)Wu, Le, Tiwari, Gulwani, Radhakrishna, Radi{\v{c}}ek, Soares, Wang, Li, and Xie}]{wu2022nl2viz}
Zhengkai Wu, Vu~Le, Ashish Tiwari, Sumit Gulwani, Arjun Radhakrishna, Ivan Radi{\v{c}}ek, Gustavo Soares, Xinyu Wang, Zhenwen Li, and Tao Xie. 2022.
\newblock Nl2viz: natural language to visualization via constrained syntax-guided synthesis.
\newblock In \emph{Proceedings of the 30th ACM Joint European Software Engineering Conference and Symposium on the Foundations of Software Engineering}, pages 972--983.

\bibitem[{Yang et~al.(2025)Yang, Li, Yang, Zhang, Hui, Zheng, Yu, Gao, Huang, Lv et~al.}]{qwen3}
An~Yang, Anfeng Li, Baosong Yang, Beichen Zhang, Binyuan Hui, Bo~Zheng, Bowen Yu, Chang Gao, Chengen Huang, Chenxu Lv, and 1 others. 2025.
\newblock Qwen3 technical report.
\newblock \emph{arXiv preprint arXiv:2505.09388}.

\bibitem[{Yang et~al.(2024{\natexlab{a}})Yang, Liu, Wang, and Liu}]{yang2024foundation}
Weikai Yang, Mengchen Liu, Zheng Wang, and Shixia Liu. 2024{\natexlab{a}}.
\newblock Foundation models meet visualizations: Challenges and opportunities.
\newblock \emph{Computational Visual Media}, 10(3):399--424.

\bibitem[{Yang et~al.(2024{\natexlab{b}})Yang, Zhou, Wang, Cong, Han, Yan, Liu, Tan, Liu, Yu et~al.}]{yang2024matplotagent}
Zhiyu Yang, Zihan Zhou, Shuo Wang, Xin Cong, Xu~Han, Yukun Yan, Zhenghao Liu, Zhixing Tan, Pengyuan Liu, Dong Yu, and 1 others. 2024{\natexlab{b}}.
\newblock Matplotagent: Method and evaluation for llm-based agentic scientific data visualization.
\newblock In \emph{Findings of the Association for Computational Linguistics ACL 2024}, pages 11789--11804.

\bibitem[{Ye et~al.(2024)Ye, Hao, Hou, Wang, Xiao, Luo, and Zeng}]{ye2024generative}
Yilin Ye, Jianing Hao, Yihan Hou, Zhan Wang, Shishi Xiao, Yuyu Luo, and Wei Zeng. 2024.
\newblock Generative ai for visualization: State of the art and future directions.
\newblock \emph{Visual Informatics}, 8(2):43--66.

\bibitem[{Zhang and Elhamod(2025)}]{d2d}
Ran Zhang and Mohannad Elhamod. 2025.
\newblock Data-to-dashboard: Multi-agent llm framework for insightful visualization in enterprise analytics.
\newblock \emph{arXiv preprint arXiv:2505.23695}.

\bibitem[{Zhang et~al.(2025)Zhang, Zhang, Wei, Xu, and You}]{zhang2025postergen}
Zhilin Zhang, Xiang Zhang, Jiaqi Wei, Yiwei Xu, and Chenyu You. 2025.
\newblock Postergen: Aesthetic-aware paper-to-poster generation via multi-agent llms.
\newblock \emph{arXiv preprint arXiv:2508.17188}.

\bibitem[{Zheng et~al.(2025)Zheng, Guan, Kong, Zheng, Zhou, Lin, Lu, He, Han, and Sun}]{zheng2025pptagent}
Hao Zheng, Xinyan Guan, Hao Kong, Jia Zheng, Weixiang Zhou, Hongyu Lin, Yaojie Lu, Ben He, Xianpei Han, and Le~Sun. 2025.
\newblock Pptagent: Generating and evaluating presentations beyond text-to-slides.
\newblock \emph{arXiv preprint arXiv:2501.03936}.

\end{thebibliography}

\clearpage
\appendix

\onecolumn
\startcontents[appendix]

\printcontents[appendix]{l}{1}{\section*{Appendix Contents}}

\section{Proof of theoretical analysis}
In this section, we provide a proof supporting the claim that our decoupled framework maximizes the reliability of dashboard generation. We formulate the problem using concepts from Information Theory, specifically focusing on \textbf{Mutual Information} and \textbf{Fano's Inequality}.
\subsection{Problem Formulation}
Let $\mathcal{I}$ denote the user's prompt and $\mathcal{V}$ denote the final generated dashboard. The goal of the generation system is to maximize the \textbf{Mutual Information (MI)} between the intent and the result:
\begin{equation}
\max I(\mathcal{I}; \mathcal{V}) = H(\mathcal{I}) - H(\mathcal{I} | \mathcal{V})
\end{equation}
Since the source entropy $H(\mathcal{I})$ is constant for a given prompt, maximizing MI is equivalent to minimizing the \textbf{Conditional Entropy} $H(\mathcal{I} | \mathcal{V})$ (information loss).
\subsection{Linking Entropy to Error Probability}
We assume the LLM generates a sequence $Y$ to construct $\mathcal{V}$. The generation process is stochastic. Let $P_e$ be the probability of error in the generation process (i.e., the generated $Y$ fails to functionally or semantically represent $\mathcal{I}$). As derived in Section~\ref{sec:theory}, the success probability $P_{succ} = 1 - P_e$ is governed by the total entropy of the target sequence $H(Y)$:
\begin{equation}
P_{succ} \propto e^{-H(Y)} = e^{-(H_{ir} + H_{vis})}
\end{equation}
Comparing \textit{NL2Dashboard} ($Y_{ours}$) with the End-to-End baseline ($Y_{base}$):
\begin{itemize}
\item \textbf{Baseline}: The target sequence includes tokens for visual presentation. Thus, $H_{vis} \gg 0$, leading to a high total entropy and a lower success probability $P_{succ}^{base}$.
\item \textbf{Ours}: The target sequence is much conciser.Visual handling is offloaded to a deterministic engine ($H_{vis} \approx 0$). This yields $H(Y_{ours}) \approx H_{ir} < H(Y_{base})$, implying $P_{succ}^{ours} > P_{succ}^{base}$, and consequently, $P_e^{ours} < P_e^{base}$.
\end{itemize}
\subsection{Proof via Fano's Inequality}
To link the error probability $P_e$ back to our objective $I(\mathcal{I}; \mathcal{V})$, we invoke \textbf{Fano's Inequality}. This fundamental theorem in information theory provides a lower bound on the error probability based on the conditional entropy, or conversely, an upper bound on the conditional entropy based on the error probability:
\begin{lemma}[Fano's Inequality]
For any estimator $\mathcal{V}$ of $\mathcal{I}$ with error probability $P_e = P(\mathcal{V} \neq \mathcal{I})$, the conditional entropy is bounded by:\begin{equation}H(\mathcal{I} | \mathcal{V}) \leq H_b(P_e) + P_e \log(|\mathcal{I}| - 1)
\end{equation}
where $H_b(P_e)$ is the binary entropy function.
\end{lemma}
The inequality implies that the upper bound of information loss $H(\mathcal{I} | \mathcal{V})$ is a monotonically increasing function of the error probability $P_e$ (for $P_e < 0.5$).

\textbf{Conclusion of Proof:}
\begin{enumerate}
\item We established that our decoupled approach strictly reduces the generation error probability: $P_e^{ours} < P_e^{base}$.
\item Applying Fano's Inequality, a lower $P_e$ necessitates a lower upper bound on the information loss $H(\mathcal{I} | \mathcal{V})$.
\item Since $I(\mathcal{I}; \mathcal{V}) = H(\mathcal{I}) - H(\mathcal{I} | \mathcal{V})$, minimizing information loss is equivalent to maximizing Mutual Information.
\end{enumerate}
Therefore, by minimizing visual entropy $H_{vis}$, \textit{NL2Dashboard} theoretically guarantees a higher Mutual Information between user intent and the visualized result.

\section{Modification Error Analysis}
As an extension of Section~\ref{sect::case_study}, we provide a detailed breakdown of the common error types and their frequency distribution in baseline models when editing existing dashboards, and summarize the result in Table~\ref{tab::error_distribution}. 
\begin{table}
\centering
\caption{Modification Error Distribution}
\renewcommand{\arraystretch}{1.3} 
\begin{tabularx}{\linewidth}{l X c} 
\toprule
\textbf{Error Type} & \textbf{Description} & \textbf{Prop.} \\ 
\midrule
Spatial Layout Error & The model fails to correctly process the spatial relationships of charts, manifested as unsuccessful position swaps, reversed left/right or up/down order, or placing a newly added chart in an incorrect coordinate. & 41\% \\
Task Omission Error & The model ignores the core generation or addition task, although it may have completed partial instructions (e.g., renaming, swapping). & 21\% \\
Over-Deletion Error & The model exhibits poor boundary control during deletion or replacement commands, leading to the unintended removal of non-target charts or even clearing the entire canvas. & 18\% \\
Intent Error & The model confuses the user's operation command types (especially "add" vs. "replace") or fails to comprehend the sequential logic of the operations. & 12\% \\
Ineffective Exec. Error & The model hallucinates, claiming task completion without actual change, or generating ineffective content (e.g., empty charts). & 8\% \\ 
\bottomrule
\end{tabularx}
\label{tab::error_distribution}
\end{table}

\section{Study of the Critic Agent}

\begin{figure}
    \centering
    \includegraphics[width=0.6\linewidth]{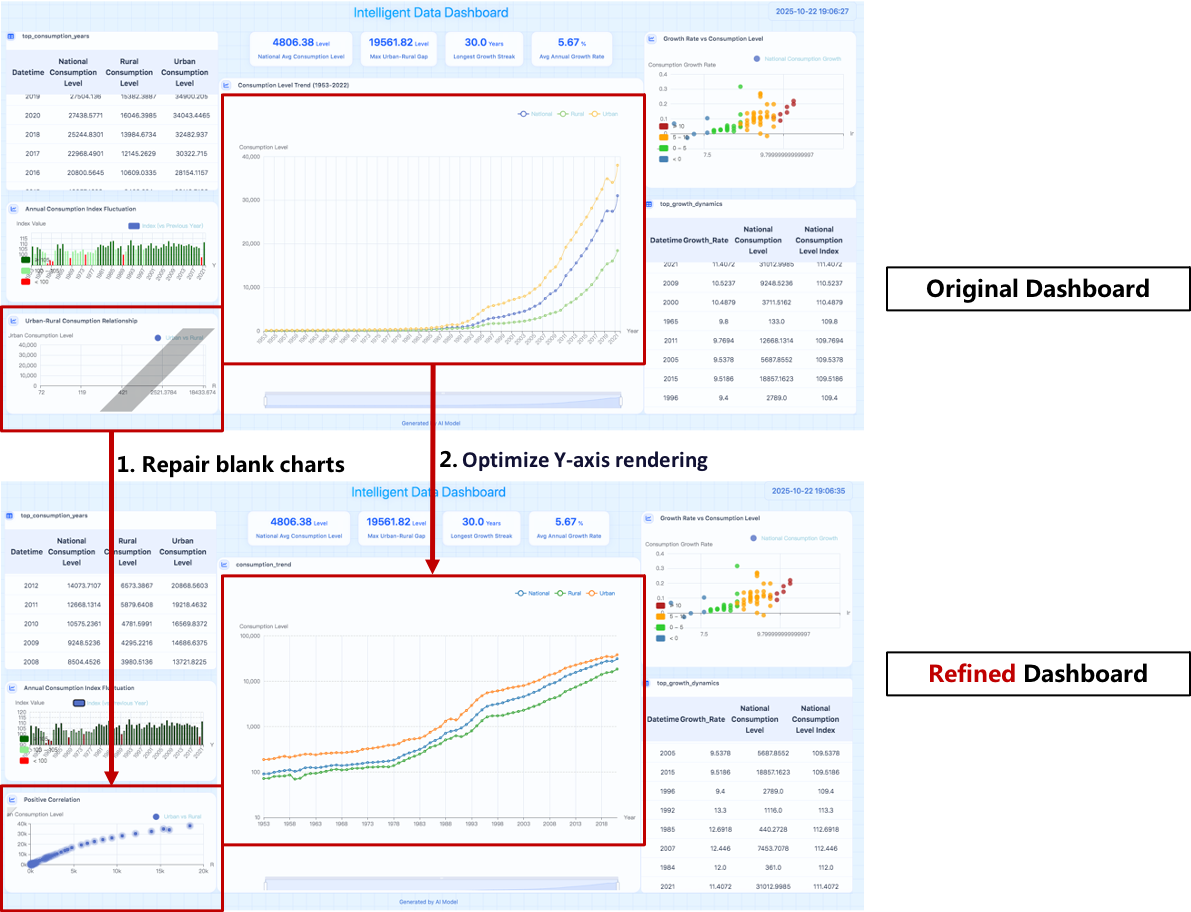}
    \caption{Critic performance}
    \label{fig::critic}
\end{figure}

As the critic agent contributes to the improvement, we carefully explored the changes brought by the critics. As illustrated in Fig.~\ref{fig::critic}, the critic agent not only repairs errors like blank charts, but also improves the visual performance.
\section{Prompts}
\label{sec:appendix}
In this section we show prompts for dashboard intent detection, dashboard generation, dashboard modification, VLM-as-a-Judge for computing dashboard qualities, and VLM critic.

\begin{tcblisting}{
    colframe=black,
    colback=white,
    coltitle=white,
    title=\textbf{1. Dashboard Intent Detection},
    sharp corners,
    boxrule=1pt,
    width=\linewidth,
    breakable,        % 【关键】允许跨页
    listing only,     % 【关键】告诉它这只是代码展示，不要运行
    listing options={
        basicstyle=\ttfamily\small,
        breaklines=true, % 允许代码行自动换行
        columns=fullflexible,
        breakindent=0pt,       % 【关键1】强制折行后的缩进为 0
        breakautoindent=false  % 【关键2】关闭自动对其上一行缩进
    }
}
### Goal
According to the user's query, please determine whether the user wants to "generate a new dashboard" or "modify an existing dashboard."

### Output Requirement
When outputting, you may only output one of the following two words: "generation" or "modify":
- "generation" means creating a new dashboard from scratch.
- "modify" means making changes to an existing dashboard.

Important: The output word must be wrapped in a <result></result> block.
Example: <result>generation</result> or <result>modify</result>

The use query is: {{USER_QUERY}}
\end{tcblisting}

\begin{tcblisting}{
    colframe=black,
    colback=white,
    coltitle=white,
    title=\textbf{2. Dashboard Generation},
    sharp corners,
    boxrule=1pt,
    width=\linewidth,
    breakable,        % 【关键】允许跨页
    listing only,     % 【关键】告诉它这只是代码展示，不要运行
    listing options={
        basicstyle=\ttfamily\small,
        breaklines=true, % 允许代码行自动换行
        columns=fullflexible,
        breakindent=0pt,       % 【关键1】强制折行后的缩进为 0
        breakautoindent=false  % 【关键2】关闭自动对其上一行缩进
    }
}
The input table is: {{USER_TABLE}}
The use query is: {{USER_QUERY}}

### Goal
The dashboard must include three complementary types of analytical outputs: charts, tables, and statistical metrics. At the end, use Python's `print` function to strictly output a list of Python-style filenames for all result files, wrapped within <result> and </result>. Once this list is printed, the task ends-no further output should follow.  
If the table contains missing values and you need to use those rows or columns, perform missing value imputation first before conducting any data analysis.

### Expert Design  
You are an expert in the domain relevant to the provided table. After reading the basic information of the table, you must first consider what key data analysis tasks are typically prioritized in this field, and whether there are common industry practices or case studies to reference. This domain knowledge forms the foundation for subsequent task planning.

### Execution Workflow  
Follow the three-phase sequence below, leveraging your expert knowledge to complete each type of task step by step.

#### Phase 1: Generate Charts  
1. Task Planning  
   - Plan 4 independent data analysis tasks, each producing a visualization chart (e.g., bar chart, line chart, pie chart, etc.). Ensure chart types are diverse, with no repeated chart type. Do not write code at this stage, and avoid overthinking.  
   - Charts should reveal insights across different dimensions (e.g., time trends, category proportions, regional distributions), and each chart should contain rich information.  
   - Preprocessing steps: 1) Impute missing values if involved; 2) Perform preliminary aggregation if the task is complex.

2. Chart Requirements  
   - Background: Must be transparent-no other background color allowed.  
   - Labels: Disable data point labels. Use control statements like: ```label_opts=opts.LabelOpts(is_show=False)```  
   - Legend: Place legend at the top or upper-right corner, with light blue font color.  
   - Content: Maximize information richness and complexity in each chart.  
   - Output Format: Save each chart as a separate HTML file (.html).

3. Code Standards  
   - pyecharts Syntax: 1) If the x-axis involves time, ensure all elements are of type `str`. For a column `x`, convert using `df['x'].astype(str)` before plotting.  
   - Data Integrity: Do not fabricate data-use only the user-uploaded table as the data source. If data is unsuitable for a chart type, switch to another.  
   - Variable Naming: Chart variable named `X`; filename as `P.html` (choose meaningful `P`, e.g., `sales_trend.html`).  
   - Render Command: Final line must be `X.render("P.html")`.

#### Phase 2: Generate Tables  
1. Task Planning  
   - Plan 2 independent data analysis tasks, each producing a sorted Top-K table (e.g., ranked by sales, growth rate).  
   - Tables should present aggregated information across different dimensions (e.g., product category, region, time period).  
   - Preprocessing steps: 1) Impute missing values if involved; 2) Perform preliminary aggregation if the task is complex.

2. Table Specifications  
   - Format: Pandas DataFrame (without row index).  
   - Row Limit: Each table must have at least 10 rows and 3 to 5 columns.  
   - Output Format: Save each table as a separate CSV file (.csv).

3. Code Standards  
   - Variable Naming: Table variable named `X`; filename as `T.csv` (choose meaningful `T`, e.g., `top_10_sales.csv`).  
   - Output Command: Final line must be `X.to_csv("T.csv")`-do not merge code for multiple tables.

#### Phase 3: Generate Statistical Metrics  
1. Task Planning  
   - Plan 4 independent statistical metric tasks, each computing a single quantitative value (e.g., total, mean, max, min, percentage). The result must be a concrete number (integer or float) with a unit.  
   - Example format: "Total GDP of Beijing is 20000 ten thousand yuan."  
   - Preprocessing steps: 1) Impute missing values if involved; 2) Perform preliminary aggregation if the task is complex.

2. Output Format  
   - Represent each metric as a dictionary with three fields:  
     {
       "Indicator": "Beijing GDP Total", 
       "Value": "20000",
       "Unit": "ten thousand yuan"
     }, where the "Indicator" description should be concise.
   - Combine all dictionaries into a list, assigned to a meaningful variable name (e.g., `city_economic_indicators`).

3. File Saving  
   - Use `json.dump` to save the list as a JSON file, with filename matching the variable name (e.g., `city_economic_indicators.json`).  
   - Encoding & Formatting: `ensure_ascii=False`, `indent=4`.

### Final Output Requirement  
1. Filename List: Compile all result filenames (.html, .csv, .json) into a list, wrapped within <result> and </result>.

### Prohibited Actions  
- Combining code for multiple task types (e.g., writing chart, table, and stats code together).  
- Using meaningless variable names (e.g., `X`, `S`).  
- Incorrect result filenames (e.g., missing extensions or non-standard naming).

### Example Output Format  
If the result is: <result>["sales_trend.html", "top_10_sales.csv", "city_economic_indicators.json"]</result>  
Then the last line of code must be:  
`print('<result>["sales_trend.html", "top_10_sales.csv", "city_economic_indicators.json"]</result>')`. Do not use `print` anywhere else in the code.

\end{tcblisting}

\begin{tcblisting}{
    colframe=black,
    colback=white,
    coltitle=white,
    title=\textbf{3. Dashboard Modification},
    sharp corners,
    boxrule=1pt,
    width=\linewidth,
    breakable,        % 【关键】允许跨页
    listing only,     % 【关键】告诉它这只是代码展示，不要运行
    listing options={
        basicstyle=\ttfamily\small,
        breaklines=true, % 允许代码行自动换行
        columns=fullflexible,
        breakindent=0pt,       % 【关键1】强制折行后的缩进为 0
        breakautoindent=false  % 【关键2】关闭自动对其上一行缩进
    }
}
The input table is: {{USER_TABLE}}
The use query is: {{USER_QUERY}}

### Guidelines
To implement modifications to the dashboard, you must strictly output a JSON-formatted modification operation list exactly once, following the exact sequence of operations specified by the user. This JSON content must be wrapped as follows: starting with ```json and ending with ```. The JSON must contain no extra content such as comments. It must be a list, where each element is a dictionary representing one modification operation on the dashboard.  

The first key in each dictionary must be "option", and its value must be one of the four actions: "change", "delete", "add", or "swap".  
The second key must be "changes", and its value depends on the specific action as follows:

#### change action
When the user wants to modify non-chart and non-table content-such as title, color scheme, etc., use the "change" action, i.e., the first field must be {"option":"change"}.  
The value of the "changes" field is a list. Each element is a dictionary where the key is the field to be modified (named exactly as in the current configuration JSON file), and the value is the new value according to the user's request.  
Note: Fields not mentioned by the user and unchanged from the original configuration JSON should NOT appear in the output JSON.
#### delete action
When the user wants to delete a chart or table, use the "delete" action, i.e., the first field must be {"option":"delete"}.  
The value of the "changes" field is a list. Each element is a dictionary indicating the position of the component to be deleted.
#### swap action
When the user wants to swap the positions of two charts or tables, use the "swap" action, i.e., the first field must be {"option":"swap"}.  
The value of the "changes" field is a list containing exactly two elements, each representing the position of one of the two components to be swapped. Each element is a position dictionary.  
Note: Both positions being swapped must already exist in the current configuration file-i.e., charts/tables must already be present at those positions.
#### add action
When the user wants to replace an existing chart/table with a new one, or add a new chart/table to the page, use the "add" action, i.e., the first field must be {"option":"add"}.  
The value of the "changes" field is a list. Each element is a dictionary indicating the position where a new chart or table will be added.

##### Layout Requirement
You must represent the position of a chart or table using the following format:  
- "position" indicates left, middle, or right on the screen.  
- "order" indicates top (1), middle (2), or bottom (3).  
The "position" can only be one of ("left", "middle", "right"), and "order" can only be one of (1, 2, 3).

For example:  
- {"position":"left","order":1} means the top-left component.  
- {"position":"right","order":3} means the bottom-right component.  


##### Special note
For "add" operations, you must NOT output the JSON-formatted modification operation list first. Instead, you must first generate one or more new charts or tables by outputting a Python code block:

###### If a new chart is needed:  
- After plotting, assign it a meaningful name (assume it's P; no path allowed). If the chart variable is X, the last line of the chart-related code block must be `X.render("P.html")`. Do not use any other output method, and do not assign a return value to this line.  
- Configuration details: The legend font color must be set to "#00E5FF"; the entire chart background color must be "transparent".

###### If a new table is needed:  
- Each resulting table must be a pandas DataFrame (without row index), and saved using `.to_csv()`. If the table variable is X, assign it a meaningful name (assume it's T; no path allowed), and the last line must be `X.to_csv("T.csv")`.  
- Each resulting table must have at least 10 rows and no more than 5 columns.

### Output Requirement
- After generating all new charts/tables in the Python code block, you must use Python print function on the very last line to output a list of filenames for all newly generated result files, strictly formatted as a Python-style list, wrapped within <result> and </result>. Example: `print('<result>["sales_trend.html", "top_10_sales.csv"]</result>')`  
- Each element must be the filename of a newly generated analytical result (charts use .html suffix, tables use .csv suffix).  
- The number of elements must exactly match the number of requested new components.  
- The type of each element (chart or table) must correspond precisely to the user's request.  
- The order of elements must exactly match the order of positions listed in the "changes" field of the "add" operations in the JSON modification list.

### Global Notes:  
1. When making modifications, do not alter any content not explicitly specified by the user.  
2. If the user's request conflicts with the current configuration template, always follow the actual layout defined in the current configuration when executing the modification.  
   Example: If the user requests to delete the bottom-right component, but the current configuration shows that the "right" column only has up to order=2, then the bottom-right component is at {"position":"right","order":2}.  
3. The JSON modification operation list must reflect the true sequence of the user's operations. Each "add" or "delete" action must involve only one chart/table per operation. Do not merge two operations of the same type if another operation occurs between them, as intermediate changes may affect the global state.

### Examples:  
#### A.  
User: Change the title to '2024 Financial Report' and the footnote to '2024', and also delete the second chart on the right.  

You must output:  
```json
[
  {
    "option":"change",
    "changes":[{"title":"2024 Financial Report"},{"footnote": "2024"}]
  },
  {
    "option":"delete",
    "changes":[{"position":"right","order":2}]
  }
]
```

#### B.  
User: Replace the top-right chart with a new chart, replace the middle-right table with a new table, then swap the middle chart with the bottom-left table.

You must output two parts:  
Part 1:  
A Python code block that generates the new chart and new table, with the last line being:  
`print('<result>["new_chart.html","new_table.csv"]</result>')`  

Part 2:  
```json
[
  {
    "option":"add",
    "changes":[{"position":"right","order":1},{"position":"right","order":2}]
  },
  {
    "option":"swap",
    "changes":[{"position":"middle","order":2},{"position":"left","order":3}]
  }
]
```

#### C.  
User: Add a new table at the bottom-right, then swap the bottom-right table with the middle chart, and finally replace the middle table with a new chart.

You must output two parts:  
Part 1:  
A Python code block that generates two new components (one table and one chart), with the last line being:  
`print('<result>["new_chart1.html","new_chart2.html"]</result>')`  

Part 2:  
```json
[
  {
    "option":"add",
    "changes":[{"position":"right","order":3}]
  },
  {
    "option":"swap",
    "changes":[{"position":"middle","order":2},{"position":"right","order":3}]
  },
  {
    "option":"add",
    "changes":[{"position":"middle","order":2}]
  }
]
```
Note: The two "add" operations cannot be merged because a "swap" operation occurs between them.

The current visualization dashboard / dashboard configuration file is:  
{{DBCONFIG}}
\end{tcblisting}

\begin{tcblisting}{
    colframe=black,
    colback=white,
    coltitle=white,
    title=\textbf{4. VLM-as-a-Judge},
    sharp corners,
    boxrule=1pt,
    width=\linewidth,
    breakable,        % 【关键】允许跨页
    listing only,     % 【关键】告诉它这只是代码展示，不要运行
    listing options={
        basicstyle=\ttfamily\small,
        breaklines=true, % 允许代码行自动换行
        columns=fullflexible,
        breakindent=0pt,       % 【关键1】强制折行后的缩进为 0
        breakautoindent=false  % 【关键2】关闭自动对其上一行缩进
    }
}
### Goal
You are an expert in data visualization with a strong background in both academic and practical perspectives. Your task is to evaluate the information delivery capability and visual quality of images based on rigorous, specific, and content-based criteria from the actual image provided by users. When in doubt, always choose the lower score.

The uploaded image is a screenshot of a dashboard html page. Please carefully analyze and independently assess it according to the following three dimensions. Each dimension should be rated on a 5-point scale (1=Poor, 5=Excellent), accompanied by detailed explanations highlighting specific visual and content evidence supporting your score.

#### **1. Insightful Depth**
Evaluates whether the dashboard reveals deep, non-obvious patterns, trends, or relationships within the data, rather than merely displaying raw data or surface-level phenomena.
- **Score 5 (Profound)**: The chart ingeniously reveals hidden trends, outliers, causations, cross-variable interactions, or integrates multi-dimensional insights.
- **Score 4 (Quite Deep)**: Includes some level of derived conclusions such as trend forecasting or significant difference annotations.
- **Score 3 (Average)**: Presents basic statistical results without further interpretation or mining.
- **Score 2 (Shallow)**: Information remains at the data table level, lacking any analytical processing.
- **Score 1 (Invalid)**: No meaningful conclusion can be drawn, or misleading expressions obscure the truth.

#### **2. Quality**
Assesses the overall presentation quality of the dashboard in a browser environment, focusing on visual clarity, layout rationality, component rendering stability, color readability, and the absence of frontend technical flaws. Emphasizes professionalism and usability as an "interactive information interface".
- **Score 5 (Excellent)**: Outstanding design in all aspects; no charts are empty. Harmonious colors, clear layouts, distinct information hierarchy, concise and readable charts.
- **Score 4 (Good)**: Generally effective design but needs minor improvements; up to one empty chart may exist. Slightly monotonous layout but does not hinder core functionality.
- **Score 3 (Moderate)**: Usable design with noticeable defects; up to two empty charts. Colors may not match well, layout somewhat messy, average readability.
- **Score 2 (Poor)**: Significant issues; most content is empty. Distracting colors, cluttered layout, unclear information hierarchy, difficult data interpretation.
- **Score 1 (Very Poor)**: Completely failed design; uncomfortable visually, unrecognizable information, chaotic layout.

#### **3. Richness**
Measures the adequacy of information conveyed by the dashboard, including multiple dimensions, layers, or contextual information.
- **Score 5 (Extremely Rich)**: Includes different forms of data representation like charts, tables, and statistical values. Charts are diverse in type, and tables have many columns.
- **Score 4 (Fairly Rich)**: Involves at least two types among charts, tables, and statistical values, with at least two types of charts.
- **Score 3 (Moderate)**: Only includes charts, missing tables and statistical values.
- **Score 2 (Sparse)**: Extremely limited information, possibly only one or two visualization results.
- **Score 1 (Empty)**: Hardly conveys any effective information.

### Output Format Requirement:
Please return the evaluation results in a **JSON array**, containing three objects corresponding to each assessment dimension. Fields include "metric" (dimension name), "explanation" (detailed explanation), and "score"(integer score from 1 to 5).
Example output structure:

```json
[
  {
    "metric": "Insightful Depth",
    "explanation": "The chart clearly illustrates long-term growth trends and annual cyclical fluctuations through the introduction of moving averages and seasonal decomposition components...",
    "score": 5
  },
  {
    "metric": "Quality",
    "explanation": "Overall clarity is good, font sizes are appropriate, yet there's slight overlap in Y-axis labels which might affect readability when zoomed out...",
    "score": 4
  },
  {
    "metric": "Richness",
    "explanation": "The chart simultaneously displays original observations, prediction intervals, and actual deviations, enriched with external event annotations, significantly enhancing informational density...",
    "score": 5
  }
]
```
\end{tcblisting}

\begin{tcblisting}{
    colframe=black,
    colback=white,
    coltitle=white,
    title=\textbf{5. VLM Critic},
    sharp corners,
    boxrule=1pt,
    width=\linewidth,
    breakable,        % 【关键】允许跨页
    listing only,     % 【关键】告诉它这只是代码展示，不要运行
    listing options={
        basicstyle=\ttfamily\small,
        breaklines=true, % 允许代码行自动换行
        columns=fullflexible,
        breakindent=0pt,       % 【关键1】强制折行后的缩进为 0
        breakautoindent=false  % 【关键2】关闭自动对其上一行缩进
    }
}
### Goal
Given a piece of code and an image of the current plot designed for a data visualization task, your task is to determine whether the quality of the plot could be enhanced.
- If no enhancement is needed, output `<result>NO</result>`, indicating that the image is sufficiently well-designed.
- If there is room for improvement, first output `<result>YES</result>`, followed by providing natural language instructions on how to enhance the plot to better fit into the dashboard. **Important:** Do not provide any Python code as part of your suggestions. Your evaluation should focus solely on stylistic improvements.

**Context:**
The provided image is part of a data visualization dashboard and was generated from a Python code block using PyECharts, which was rendered as HTML and then converted to a PNG image. The relevant code snippet is enclosed within `{{CODE}}`.

**Guidelines:**
- Do not suggest modifications to the original content. Specifically:
  1. Keep the background color of the image transparent at all times.
  2. Ensure the legend remains in light blue font and is positioned at the top or top-right corner of the image.
  3. Data point labels should remain hidden.
  4. Do not alter the original chart title; moreover, ensure the title is not displayed.
- If you notice that the image appears blank, this indicates an error in the preceding code that prevented content from being rendered. Address this issue as your primary recommendation.
- In the absence of significant errors or issues with the chart, refrain from suggesting changes, i.e., output `<result>NO</result>`.

Please provide detailed step-by-step instructions for any suggested enhancements.
\end{tcblisting}

\section{Base Template}
In this section we show an example of base template, in which only the most essential parts are displayed. The red-highlighted segments represent slots:
\begin{enumerate}
    \item Textual part: Slots like 'title' and 'footnote' are rendered with textual contents from IR.
    \item Analysis Component part: Slots like 'TODO-DEPENDENCE', 'TODO-LEFT-COLUMN-CONTENT', and 'TODO-JS-Chart' are rendered from analysis component files like HTML-formatted charts with Echarts.
\end{enumerate}
\begin{tcblisting}{
    colframe=black,
    colback=white,
    coltitle=white,
    title=\textbf{Base Template Example},
    sharp corners,
    boxrule=1pt,
    width=\linewidth,
    breakable,        % 【关键】允许跨页
    listing only,     % 【关键】告诉它这只是代码展示，不要运行
    listing options={
        basicstyle=\ttfamily\small,
        escapeinside={(*}{*)},
        breaklines=true, % 允许代码行自动换行
        columns=fullflexible,
        breakindent=0pt,       % 【关键1】强制折行后的缩进为 0
        breakautoindent=false  % 【关键2】关闭自动对其上一行缩进
    }
}
<!DOCTYPE html>
<html lang="zh-CN">
<html class="dark">
<head>
    <meta charset="UTF-8">
    <meta name="viewport" content="width=device-width, initial-scale=1.0">
    <title>(*\textcolor{red}{\{\{title\}\}}*)</title>
    <link href="https://cdnjs.cloudflare.com/ajax/libs/font-awesome/6.4.2/css/all.min.css" rel="stylesheet">
    <script src="https://cdn.tailwindcss.com"></script>
    (*\textcolor{red}{\{\{TODO-DEPENDENCE\}\}}*)
    <script type="text/javascript" src="https://assets.pyecharts.org/assets/v5/maps/china.js"></script>
    <style type="text/tailwindcss">
   ...
    </style>
    <style>
        html, body {
            height: 100%;
            margin: 0;
            overflow: hidden; 
        }
    ...
    </style>
</head>
<body class="grid-bg">
    <div id="dynamic-clock"></div>
    <header>
        <h1>(*\textcolor{red}{\{\{title\}\}}*)</h1>
    </header>
    <div class="dashboard-body">
        <div class="grid-col-left">
            (*\textcolor{red}{\{\{TODO-LEFT-COLUMN-CONTENT\}\}}*)
        </div>
        <div class="grid-col-middle">
            (*\textcolor{red}{\{\{TODO-MIDDLE-COLUMN-CONTENT\}\}}*)
        </div>
        <div class="grid-col-right">
            (*\textcolor{red}{\{\{TODO-RIGHT-COLUMN-CONTENT\}\}}*)
        </div>
    </div>
    <footer>
        <p>(*\textcolor{red}{\{\{footnote\}\}}*)</p>
    </footer>
    (*\textcolor{red}{\{\{TODO-JS-Chart\}\}}*)
    ...
</body>
</html>

\end{tcblisting}

\end{document}